 \documentclass[a4paper, 10pt, conference]{ieeeconf}      

\IEEEoverridecommandlockouts                    

\usepackage{algorithm}

\usepackage[utf8]{inputenc}
\usepackage[english]{babel}
\usepackage{setspace}
\usepackage{cite}
\usepackage{multirow}
\usepackage[table,xcdraw]{xcolor}
\usepackage{subfigure}
\usepackage{graphics} 
\usepackage{graphicx}
\graphicspath{{figures/}}
\usepackage[mathscr]{euscript}
\usepackage{listings}
\usepackage{color}
\usepackage{listings}

\usepackage{graphicx,wrapfig,lipsum}

\usepackage{tikz}
\usetikzlibrary{intersections}
\usetikzlibrary{decorations.pathreplacing}

\usepackage{multicol}
\usepackage{algorithm}
\usepackage[noend]{algpseudocode}

\usepackage{amsthm} 
\usepackage{amsmath} 
\usepackage{amssymb}  
\usepackage[percent]{overpic}
\usepackage{chemformula}
\usepackage{multirow}
\usepackage{array}

\theoremstyle{plain}

\definecolor{lightgray}{gray}{0.97}
\definecolor{lightgray2}{gray}{0.1}

\begin{document}


\title{Piecewise Affine Curvature model: a reduced-order model for soft robot-environment interaction beyond PCC}

\author{Francesco Stella$^{1,3*}$, Qinghua Guan$^{1,2*}$,  Jinsong Leng$^{2}$, Cosimo Della Santina$^{3,4}$, Josie Hughes$^{1}$ 
\thanks{$^{*}$ These authors contributed equally to this work. $^{1}$CREATE Lab, EPFL, Lausanne, Switzerland. $^{2}$Harbin Institute of Technology, Harbin, China. $^{3}$Department of Cognitive Robotics, Delft University of Technology, Delft, The Netherlands. $^{4}$Institute of Robotics and Mechatronics, German Aerospace Center (DLR), Wessling, Germany.
{\footnotesize Contact emails: \texttt{francesco.stella@epfl.ch, guanqinghualx@foxmail.com, lengjs@hit.edu.cn, josie.hughes@epfl.ch, c.dellasantina@tudelft.nl}}.
}
}

\maketitle
\begin{abstract}
Soft robot are celebrated for their propensity to enable compliant and complex robot-environment interactions. 
Soft robotic manipulators, or slender continuum structure robots have the potential to exploit these interactions to enable new exploration and manipulation capabilities and safe human-robot interactions. 
However, the interactions, or perturbations by external forces cause the soft structure to deform in an infinite degree of freedom (DOF) space. 
To control such system, reduced order models are needed; typically models consider piecewise sections of constant curvature although external forces often deform the structure out of the constant curvature hypothesis. 
In this work we perform an analysis of the trade-off between computational treatability and modelling accuracy. 
We then propose a new kinematic model, the \textit{Piecewise Affine Curvature} (PAC) which we validate theoretically and experimentally showing that this higher-order model better captures the configuration of a soft continuum body robot when perturbed by the external forces.  
In comparison to the current state of the art Piecewise Constant Curvature (PCC) model we demonstrate up to 30\% reduction in error for the end position of a soft continuum body robot.

\end{abstract}


\maketitle
\section{Introduction}
Soft robots have long promised to change the way robots interact for the environment. One significant challenge is the design, modeling and control of soft manipulators. They have been developed in many forms and structures \cite{rus2015design,dellaCatalanoBicchi}, and can show a wide range of deformation in response to interactions with the environment\cite{li2017kinematic}. Moreover, thanks to their compliance they inherently show robust reactions to impacts and adaptation to the environment.
However, their capacity to deform in theoretically infinite degrees of freedom opens clear modeling and control challenges \cite{della2021model}. To address these challenges, several control-oriented models of soft robots\cite{grazioso2019geometrically,naughton2021elastica,renda2020geometric,duriez2013control} have been developed so to be manageable in terms of complexity and number of states, while approximating the theoretically infinite deformation stated \cite{altenbach2013cosserat}.\\
\begin{figure}[tb]
    \centering
     \vspace{1mm}
    \includegraphics[width=1\linewidth]{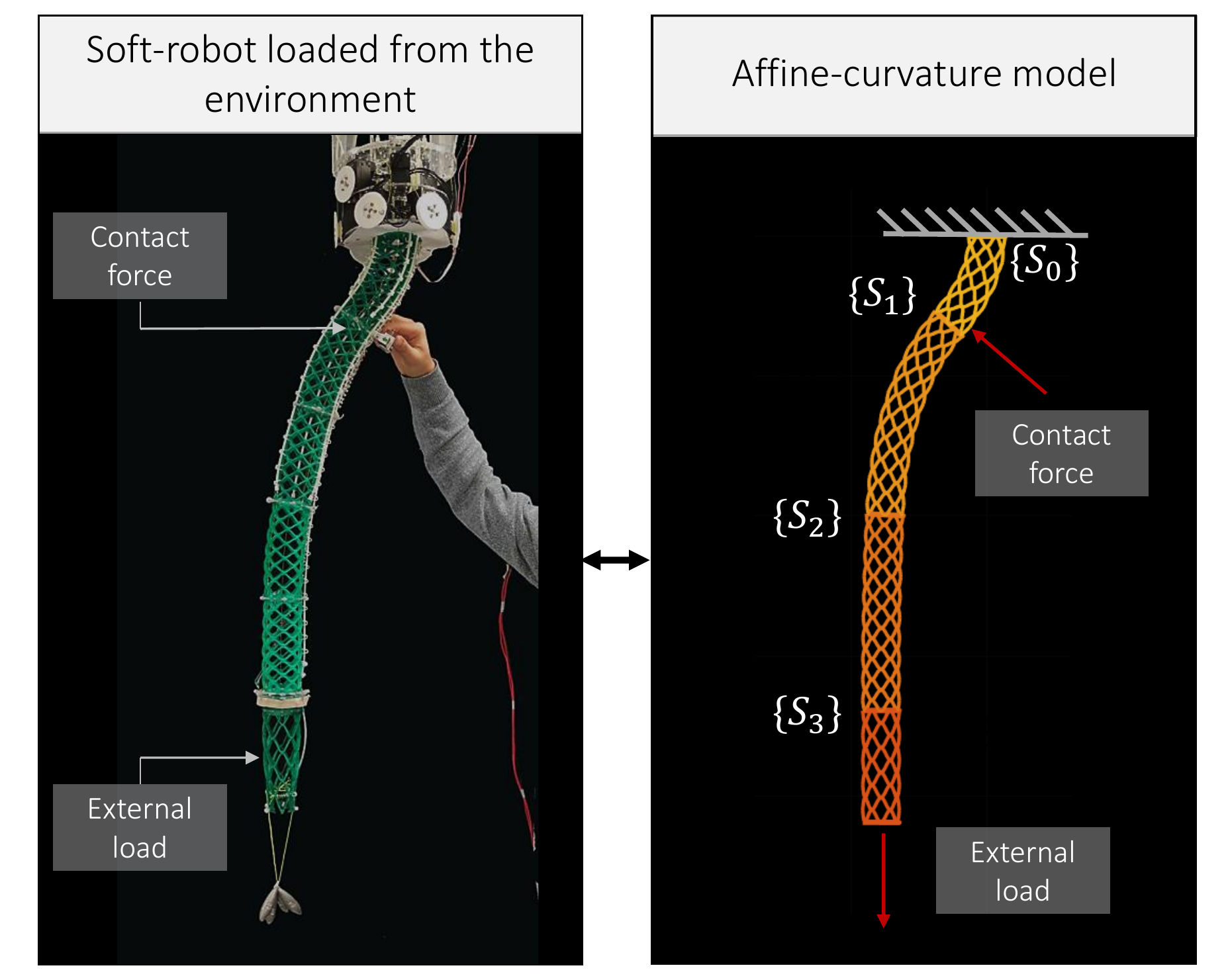}
    \caption{Pictorial figure showing the improved reconstruction capabilites of the PAC model for soft manipulators undergoing external loads. }
    \label{fig:fig1}
    \vspace{-4mm}
\end{figure}
For soft slender robots, the most commonly used method is the celebrated Piecewise Constant Curvature (PCC) model \cite{webster2010design}, which is obtained from Cosserat's rod model \cite{till2019real} by neglecting all strains but curvature and assuming that the curvature is piecewise constant in space, with fixed nodes. Such a model is theoretically exact only if the soft robot is statically perturbed with a constant bending moment along each segment and the structure is homogeneous in space \cite{webster2010design}.
Despite its simplicity, the PCC proved to well approximate both the kinematic \cite{hughes2020sensing} and dynamics \cite{della2018dynamic,Milana} of lightweight robots, composed of homogeneous materials and stiffness \cite{besselaar2022one} and subject to minimal interaction with the environment. 
However, whenever these hypotheses are not fulfilled, the soft robot will be eventually deformed out of the constant curvature hypothesis. The minimalism of the PCC model comes at the cost of a poor reconstruction when the structure interacts with the environment,  carries a load or the self-weight is not negligible. This, ultimately limits applications of soft manipulators to very confined and controlled environments. To bring soft robots toward real-world applications, it is crucial to develop models inspired by the minimalism of PCC, but able to robustly model the interaction with the environment. Such quest naturally opens the question: what is the optimal trade-off between model accuracy and computational treatability?   \\

 To address this challenge the polynomial curvature approach was recently proposed t \cite{della2019control}. In this model, the accuracy of the reconstruction can be tuned by the order of the polynomium. This has shown to provide better reconstructions of the robot configuration when the structure is not homogeneous and subject to dynamical excitations \cite{stella2022experimental}. In this work we demonstrate that the first-order truncation of the polynomial-curvature model \cite{della2020softinverted}, i.e. the affine-curvature model, significantly outperforms the accuracy of PCC models when an external load from the environment deforms the soft structure. 
In the following, first we introduce the kinematic and dynamic description of a three-dimensional \textit{Piecewise Affine-Curvature} (PAC) model. We also propose a simple strategy for solving the statics of a robot modeled this way. We then present and characterize a novel soft manipulator used as a platform to experimentally demonstrate the modeling advantages of the PAC model when the robot is perturbed by an external force or when subject to significant self weight. 


To summarize, this paper contributes to the state of the art in control-oriented models for soft robots with:
\begin{itemize}
    \item the three dimensional \textit{Piecewise Affine Curvature} (PAC) model for extensible, continuum soft robots,
    \item an experimental validation of the improved performance provided by the PAC model, with a focus on the static equilibrium in soft robot-environment interaction.
    \item a comparison between the reconstruction accuracy provided by PAC with respect to PCC models.
\end{itemize}

In the remainder of this paper we first introduce the 3D PAC model in Sec.\ref{affine_model} after which we present the soft robotics platform we use to experimentally validate the approach in Sec.\ref{setup}. The performance of the PAC model is first shown on a single section soft robot, before demonstrating in Sec.\ref{results} the ability of the PAC model to capture the effect of external forces on a three segments robot.

\section{3D Piecewise Affine Curvature model }
\label{affine_model}
\subsection{Kinematic model}
\begin{figure}[tb]
    \centering
        \vspace{1mm}
    \includegraphics[width=1\linewidth]{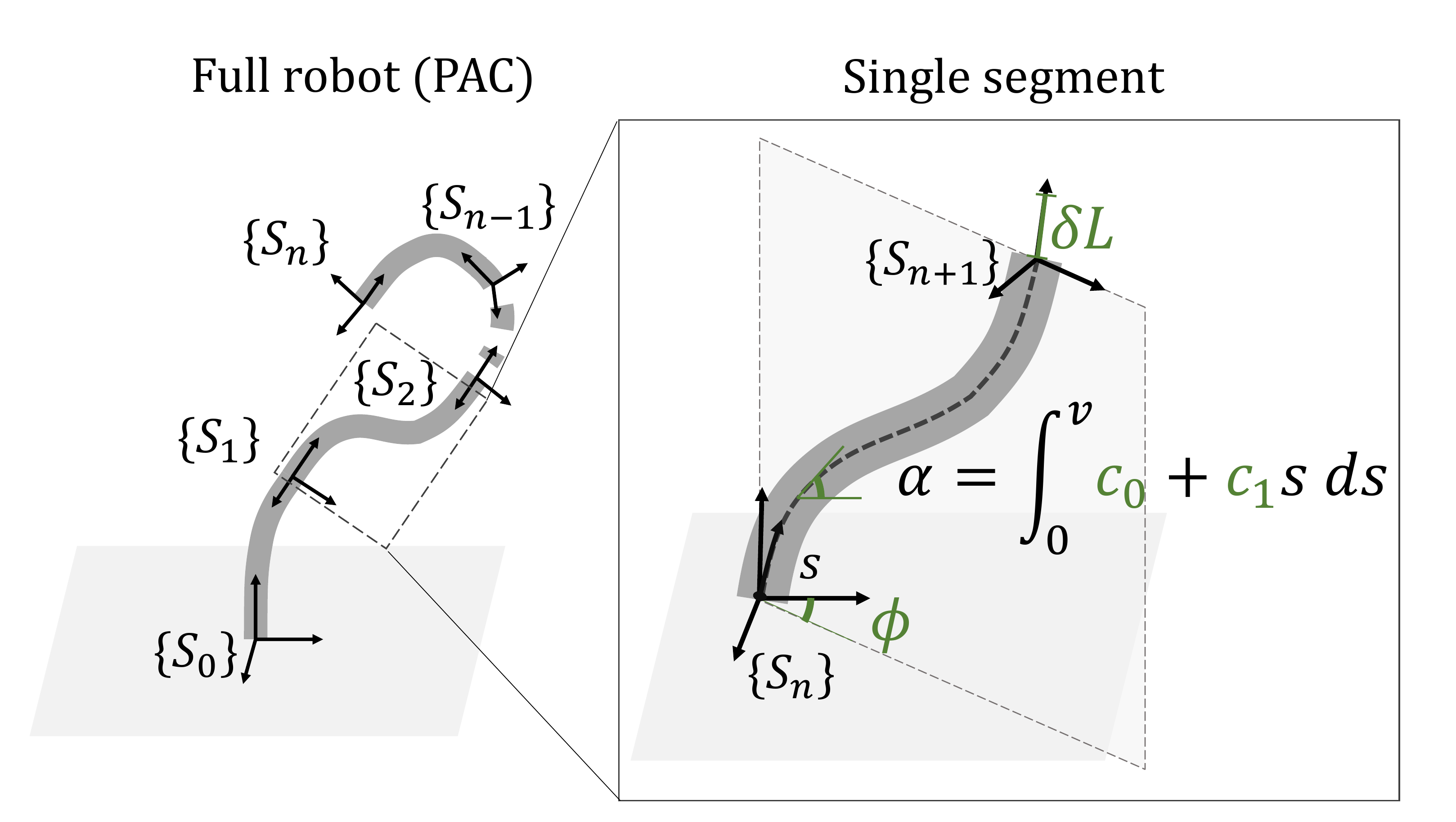}
    \caption{Schematic representation of the Piecewise Affine Curvature kinematic model. On the left, the multi segment representation, while on the right it is possible to observe a closeup on a single segment, with the the Lagrangian variables highlighted. }
    \label{fig:kinematic}
    \vspace{-4mm}
\end{figure}

We introduce here the kinematic structure of the 3D PAC model introduced in this work (see Fig. \ref{fig:kinematic}). 
This model builds upon the one-segment planar model in \cite{della2019control}, later validated experimentally for the 2D case in \cite{stella2022experimental}, extending it to the multi-segment variable-length 3D case. 

We start from the description of the curvature of its central axis shape. 
We introduce $n$ reference frames $\{S_\mathrm{1}\},..., \{S_\mathrm{n}\}$ attached at the ends of each segment, plus one fixed base frame $\{S_0\}$.

We call $T_\mathrm{i-1}^{i}$ the homogeneous transformation mapping 
\begin{equation}\small 
    T_\mathrm{i-1}^\mathrm{i}=\begin{bmatrix}
    R_\mathrm{i-1}^\mathrm{i} & t_\mathrm{i-1}^\mathrm{i}\\
    [0 \; 0 \; 0] & 1
    \end{bmatrix},
\end{equation}
with $R_\mathrm{i-1}^\mathrm{i} =
[\{n_i\}_{i-1} \; \{e_i\}_{i-1} \; \{o_i\}_{i-1}] \in
 \mathrm{SO(3)} $ rotation
matrix, and $t_\mathrm{i-1}^\mathrm{i} $ translation. $\{n_i\}, \; \{e_i\}, \; \{o_i\}$ are
orthonormal unit vectors which identify the three axes of $\{S_\mathrm{i}\}$, with
coordinates expressed w.r.t. $\{S_\mathrm{i-1}\}$.
In this novel 3D model, each segment is free to bend in any direction; its curvature is affine in space but variable in time and the segments are connected so that the resulting curve is everywhere differentiable. We consider here that each segment can also change length. 
The configuration of each segment can indeed be described by the coordinates $q_i = [c_{0,\mathrm{i}},c_{1,\mathrm{i}},\phi_\mathrm{i}, \delta L_\mathrm{i}] \in \mathbb{R}^4$ such that 
\begin{itemize}
    \item $c_{0,\mathrm{i}}$ is the zero-order term of the curvature polynomium,
    \item $c_{1,\mathrm{i}}$ is the first-order term of the curvature polynomium,
    \item $\phi_\mathrm{i}$ is the the angle between the plane created by the linear combination of $\{n_i\}_{i-1},\{e_i\}_{i-1}$ and the plane on which the bending occur,
    \item $\delta L_\mathrm{i}$ is the change in length of the central axis due to compression or extension of the segment.
    \end{itemize}
We assume that the curvature of the central axis of the $\mathrm{i}-$\textit{th} segment can be described by the affine function:
\begin{equation}\small 
    c(t)_\mathrm{i}=c_{0,\mathrm{i}}(t)+c_{1,\mathrm{i}}(t)s
\end{equation}
where $s\in [0,1]$ parameterizes the position along the main axis of the structure, so that $(L_\mathrm{i}+\delta L_\mathrm{i})s$ is the arc length of the path connecting the base to the point $s$ through the main axis.

The angle of the central axis $\alpha$ on the plane defined by $\phi$ can be found by integration of the curvature: 
\begin{equation}\small 
    \alpha(s)=\int_0^s c(\textcolor{black}{v}) \mathrm{d}\textcolor{black}{v} =c_0 s+ \frac{c_1}{2} s^2,
\end{equation}
where $v$ is an auxiliary variable with the same meaning as $s$.
The relative rotation $R_\mathrm{i-1}^\mathrm{s}$ between the $\{{S}_{i-1}\}$ and the frame $\{S_{s}\}$ attached to the $s$ point, expressed in the $\{{S}_{i-1}\}$ frame, can be written as 
\begin{equation}\small 
    R_\mathrm{i-1}^\mathrm{s} =\left[\begin{array}{ccc} \cos\left(\alpha\right)\,\cos\left(\phi \right) & -\sin\left(\phi \right) & \sin\left(\alpha\right)\,\cos\left(\phi \right)\\ \cos\left(\alpha\right)\,\sin\left(\phi \right) & \cos\left(\phi \right) & \sin\left(\alpha\right)\,\sin\left(\phi \right)\\ -\sin\left(\alpha\right) & 0 & \cos\left(\alpha\right) \end{array}\right]
\end{equation}
 while the translation vector $t_\mathrm{i-1}^\mathrm{s}$ can be written as
 \begin{equation}\small 
    t_\mathrm{i-1}^\mathrm{s} =\left[\begin{array}{ccc} \cos(\phi)\int_0^{\textcolor{black}{s}} (L+\delta L) \sin(\alpha\textcolor{black}{(v)}) \mathrm{d}\textcolor{black}{v}\\ \sin(\phi)\int_0^{\textcolor{black}{s}} (L+\delta L) \sin(\alpha\textcolor{black}{(v)}) \mathrm{d}\textcolor{black}{v}\\ \int_0^{\textcolor{black}{s}}  (L+\delta L) \cos(\alpha\textcolor{black}{(v)}) \mathrm{d}\textcolor{black}{v}\end{array}\right]
 \end{equation}
similarly, $v$ is an auxiliary variable with the same meaning as $s$. The integrals in $t_\mathrm{i-1}^\mathrm{s}$ can be expressed in closed form as combination of Fresnel integrals or Gauss error function as discussed in \cite{della2020softinverted}.

\subsection{Dynamical model}


Using Lagrangian derivations, we can build\footnote{We omit the steps for the sake of space.} the general dynamical model of the robot from the kinamtic model introduced above  \cite{della2021model} 
\begin{equation}\label{eq:full_dynamics}\small
     M(q)\Ddot{q}+C(q,\Dot{q})\dot{q}+G(q)+D(q)\Dot{q}+K(q) \!=\!A(q)f_\mathrm{Act} + J^\top(q) f_\mathrm{ext},
\end{equation}
where $M(q)$ and $C(q,\dot{q})$ represent the inertia matrix and the Coriolis terms respectively, $G(q)$ is the gravitational force, and $D$ and $K$ are the damping and stiffness forces. $J$ represents the manipulator Jacobian $J(q)=\frac{\partial f(q)}{\partial q}$, where $f(q)$ is the forward kinematics mapping, which defines the point contact between the manipulator and the environment as a function of $q$. In the following we will consider this point to be the end-effector of the robot, without losing generality.

Finally, the robot interacts with the environment through an external wrench $f_\mathrm{ext}$. For generality we consider the case in which the actuation forces $f_\mathrm{Act}$ are not directly collocated at the joints coordinates, such that there is an intermediate mapping $A(q)$ between the actuation torques $\tau_\mathrm{Act}$ and the torques on the states $\tau_\mathrm{q}$.

The experimental validation of this paper will be concerned with quasi-static equilibrium conditions $\dot{q},\ddot{q}\approx0$. The dynamics of the robot can be then simplified to
\begin{equation}\small 
\label{eq_static_eq}
    G(q)+K(q)=[A(q) \;J^\top(q)]\begin{bmatrix}f_\mathrm{Act},\\ f_\mathrm{ext}\end{bmatrix}.
\end{equation}
The gravitational force, can be expressed as: 
\begin{equation}\small 
    G=\int_0^{L+\delta L} \int_{-r(s)}^{r(s)}-\rho g z_{s,d}\textcolor{black}{(v,d)} \;  \,\mathrm{d}\textcolor{black}{v} \, \mathrm{d}d,
\end{equation}
where $g$ represents the gravitational constant, $\rho$ the density of soft robot's material and $r(s)$ the radius of the robot section in $s$.
The elastic field can be determined by differentiation of the the potential elastic energy: 
\begin{equation}\small 
    U_\mathrm{E}(q,k)=\frac{1}{2}\int_0^1 k(s)\alpha^2(s,t) \mathrm{d}s,
\end{equation}
where $k(s):[0,1]\rightarrow \mathbb{R}^+$ associates a local flexural stiffness to any point along the central axis. As introduced in \cite{della2020softinverted}, the submatrix of stiffnesses on the curvatures takes the shape of a Hankel matrix $H_\mathrm{i,j}=1/(i+j-1) \in \mathbb{R}^{2 \times 2}$, while the rest of the matrix has the diagonal elements representing the stiffness on the rotation of the plane and the axial stiffness
\begin{equation}\small 
    K_\mathrm{i}(q_\mathrm{i})=\begin{bmatrix}
    k_\mathrm{bending}H\left[\begin{array}{c} c_\mathrm{0,i},\\c_\mathrm{1,i}\end{array}\right] & \begin{array}{c} 0\\0\end{array}&\begin{array}{c} 0\\0\end{array}\\
    \begin{array}{cc} 0&0\end{array}&k_\mathrm{torsion}&0\\
    \begin{array}{cc} 0&0\end{array}&0&k_\mathrm{axial}
    \end{bmatrix}
     q_\mathrm{i}
\end{equation}

\subsection{Solution of the statics}
Eq. (\ref{eq_static_eq}) is trascendental and cannot in general be solved in closed form. Also, the solution may in general not be unique making it quite challenging to even approximate a closed form relation mapping $f_\mathrm{Act}, f_\mathrm{ext}$ into an equilibrium configuration $\bar{q}$. 
A computationally intensive alternative is to obtain the equilibrium by forward integration of the complete dynamics \eqref{eq:full_dynamics}.
Instead, we propose here to evaluate $\bar{q}$ by simulating the following simpler dynamic system, which is built so to have the same equilibria
\begin{equation}\small 
    D\dot{q}+G(q)+K(q)=[A(q) \;J^\top(q)]\begin{bmatrix}f_\mathrm{Act},\\ f_\mathrm{ext}\end{bmatrix},
\end{equation}  
where $D \in \mathbb{R}^{4 \times 4}$ represents a full-rank damping matrix, which can be taken as a costant approximation of $D(q)$.


\section{Experimental setup}
\label{setup}
\subsection{Robot design}

\begin{figure}[t]
    \centering
        \vspace{1mm}
    \includegraphics[width=1\linewidth]{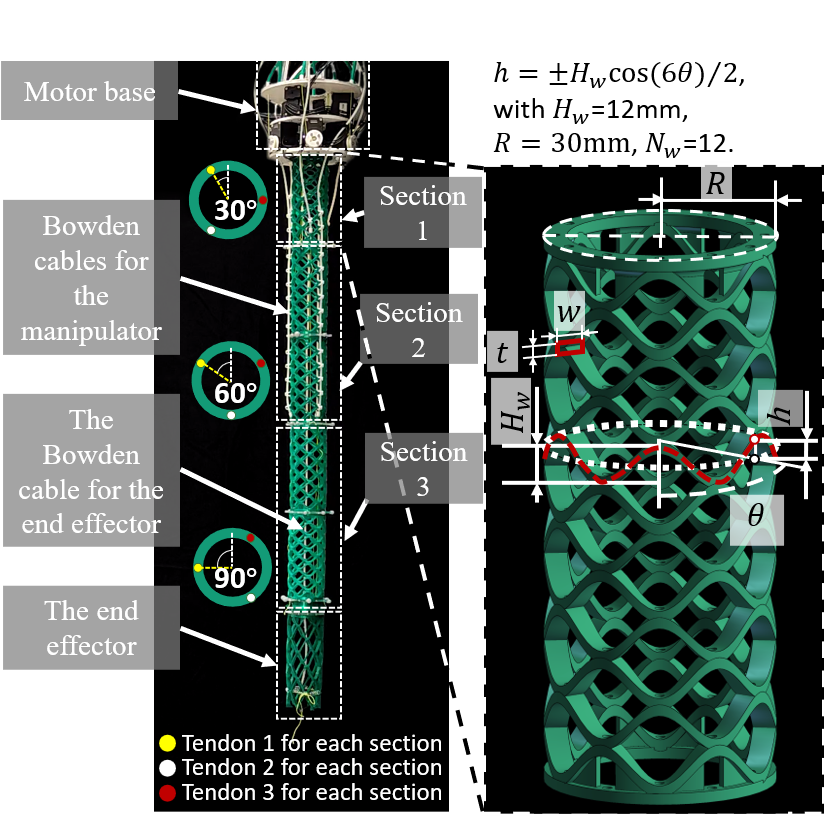}
    \caption{Schematic representation of the soft manipulator used to perform the experimental validation of this model. On the left, an image of the three segment manipulator, with the actuation methodology and singular sections highlighted. On the right, the representation of a single section and of the analytical rules defining the design.  }
    \label{fig:Robot design}
    \vspace{-4mm}
\end{figure}

We demonstrate the capabilities of the PAC model on a soft tendon-driven manipulator shown in Figure (\ref{fig:Robot design}). 
Soft manipulators are usually composed of a constant or finite-variable axis length, which limits the work space of the manipulator \cite{wand2022prismatic,li2017kinematic}.  The design of this robot arm is based on an analytically specified topology which enables high deformation ratios for both bending and contraction. These features lead to a manipulator with a large working space and which also shows non-constant curvatures under external forces. In this section we detail the design and characterization of the arm pertinent to the PAC model, however full details of the design robot can be found in~\cite{underreview}. 

The manipulator is composed of three independently actuated section placed in series, providing a redundant workspace for the complete soft arm. The manipulator is driven by ten Dynamixel XM430-W210 motors. While the first three motors are directly connected to the first section with pulleys and tendons directly, the motors actuating the second and third section are connected through Bowden cables to ensure the independent control of each section.\\
The structure of each segment of the manipulator is generated by stacking multiple sine-wave ring structures which are fabricated by FDM 3D printing with TPU. The cross section was set as a rectangle with height of 3mm and width of 6mm. Each section is composed of 12 wave rings but only ten of them are deformable, as presented in Fig. \ref{fig:fig1}. When compressed both the axial length of the structure and the radius of structure change. To minimize the magnitude of this phenomenon, the two wave rings at each end are connected to the end rings with ribs to strengthen structure.\\

\subsection{A simple model of actuation}
%

\begin{wrapfigure}{r}{0.2\textwidth}
\centering
    \includegraphics[width=0.2\textwidth]{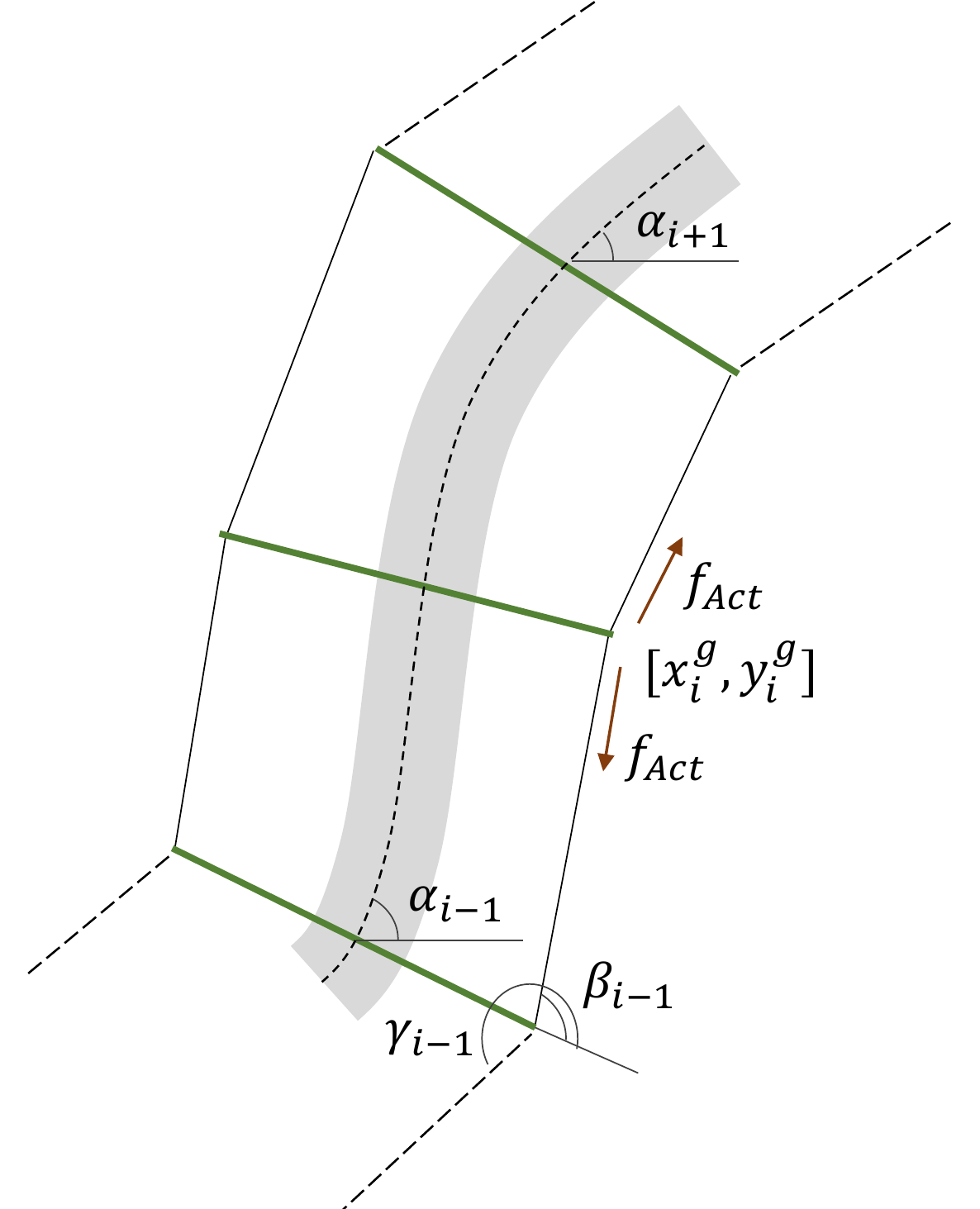}
    \caption{Schematic representation of the tendon forces on the structure, for a simplified 2D case.}
    \label{fig:tendon}
\end{wrapfigure}

We propose to derive the actuation wrench $f_\mathrm{Act}$ for such structure from the lengths of the tendons by simulating the low level control loop running in the motors.  In particular, the torque can be computed as
\begin{equation}\small 
   f_\mathrm{Act}=\kappa_{\mathrm{D}}(\dot{\bar{l}} - \dot{l}) + \kappa_{\mathrm{P}}(\bar{l} - l)),
\end{equation}
where $l,\bar{l}$ are the current and desired length of the tendon and $\kappa_{\mathrm{P}},\kappa_{\mathrm{D}}$ are two positive control gains. In the following we take both equal to $20$. Note that, as we will be working mostly under quasi-static conditions, small changes in gains will not strongly affect the performance of our model.\\

Finally, the matrix $A(q)$, which maps the actuation force input $\tau_\mathrm{Act}$ on forces on the states $f_\mathrm{q}$ is evaluated through the geometric relations in Fig.\ref{fig:tendon}, by modeling each guide-point of the tendon as a friction-less pulley. 
The force that the i$-th$ tendon apply on the j$-th$ guide, when projected on the states can be written as
\begin{equation*}\small 
 f_\mathrm{q}=J_\mathrm{g}(q)^\top \begin{bmatrix}\cos(\alpha_j) &-\sin(\alpha_j)\\ \sin(\alpha_j)&\cos(\alpha_j)\end{bmatrix}^\top
\begin{bmatrix}\cos(\beta_j) +\cos(\gamma_j)\\ 0\end{bmatrix}f^i_\mathrm{Act},
\end{equation*}

\subsection{Mechanical Characterization}

\begin{figure}[h]
    \centering
    \includegraphics[width=0.75\linewidth]{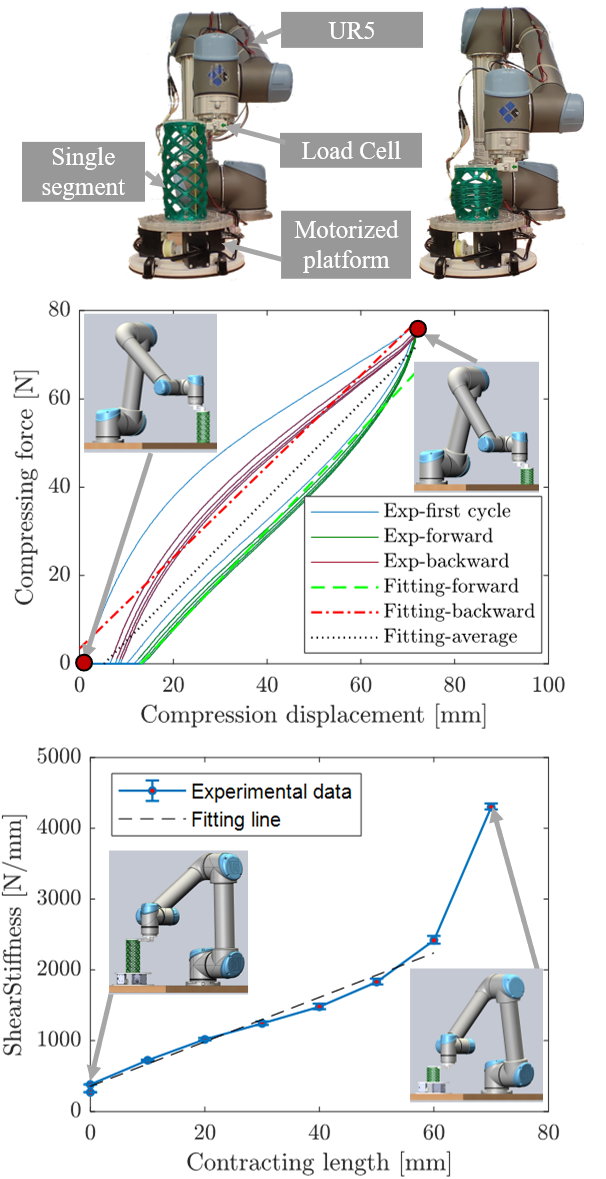}
    \caption{Experimental evaluation of the stiffness of the a single segment. The structure is compressed by a load-cell connected to a UR5 robotic manipulator. The position and force data are then post-processed to evaluate the stiffness of the structure. The experiment is repeated for several lengths, and the relation between structure length and stiffness is extracted.  }
    \label{fig:Segment test}
    \vspace{-4mm}
\end{figure}
\begin{figure}[h]
    \centering
    \includegraphics[width=0.95\linewidth]{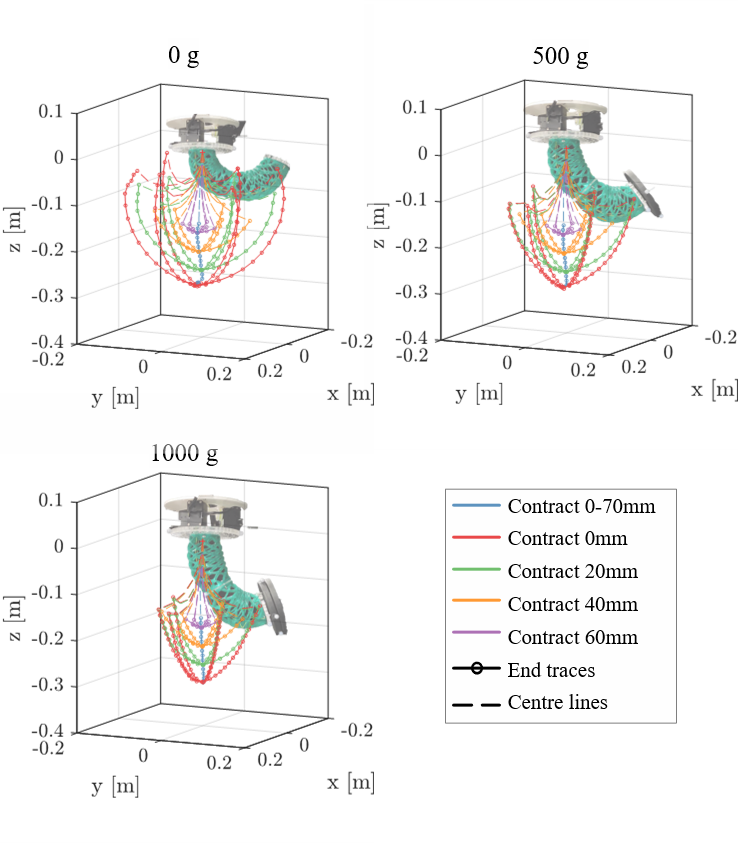}
    \caption{Analysis of the workspace as a function of the external forces. Different loads [0,500,1000]g are placed at the tip of the soft robot. Interestingly, the workspace reduces with the load, and the curvatures of the structure vary increasingly from the PCC description with increasing loads. }
    \label{fig:Single section working space}
    \vspace{-4mm}
\end{figure}
The structure of each segment was characterized with axial compressing and side shearing load tests to provide the necessary parameters for the PAC model. During the compressing test, the segment part was connected on a load cell and fasten on a UR5 arm. The UR5 arm compressed the structure from 0 to 70 mm with a speed of 5 mm/s. The stiffness resulting from the linear fitting is 1.078 N/mm. To compensate for the the visco-elastic properties of this structure, made of 3D printed TPU, the quasi-static free length was stabilized around 135.4mm after one loading cycle, which was utilized as the free length in the kinematic and static/dynamic models.\\ 
In the shear loading test, the segment was driven by three tendons connected with three Dynamixel motors. During the test, the segment contracted in different length from 0 to 70 mm under the drive of three motors, and the UR5 arm installed a load-cell compressed the segment from the side, as Figure \ref{fig:Segment test}. From Fig.\ref{fig:Segment test}, it can be seen that the stiffness of each module increases significantly with the contraction of the segment.

\subsection{Working Space Characterization}



To show the working space of the single segment, tests were conducted with a single section constructed from two segments driven by three tendons. The experiment was repeated with different loads (0, 500 and 1000g) placed on the end of the manipulator, so to observe the load-workspace interdependence. The end-effector position is tracked using motion capture (OptiTrack, with 6 cameras) for different tendon motions with the results shown in Fig.\ref{fig:Single section working space}. The results show that the working space range is significantly reduced as the load is increased. The pose and structure of the robot section also deviates increasingly from the constant curvature behaviour as the load increases, such that the PCC assumption is increasingly less valid. This characterization of the workspace indicates that to accurately model the soft robot under loads we must consider the effect of external forces, and also incorporate variable curvature into the modelling.

\section{Experimental Results}
\label{results}

In this section we validate the need of higher order models by comparing the deformation of the soft robot structure for different loading condition for the PAC and PCC models.

\subsection{Single section model evaluation}

\begin{figure}[t]
 \centering
   \includegraphics[width=1\linewidth]{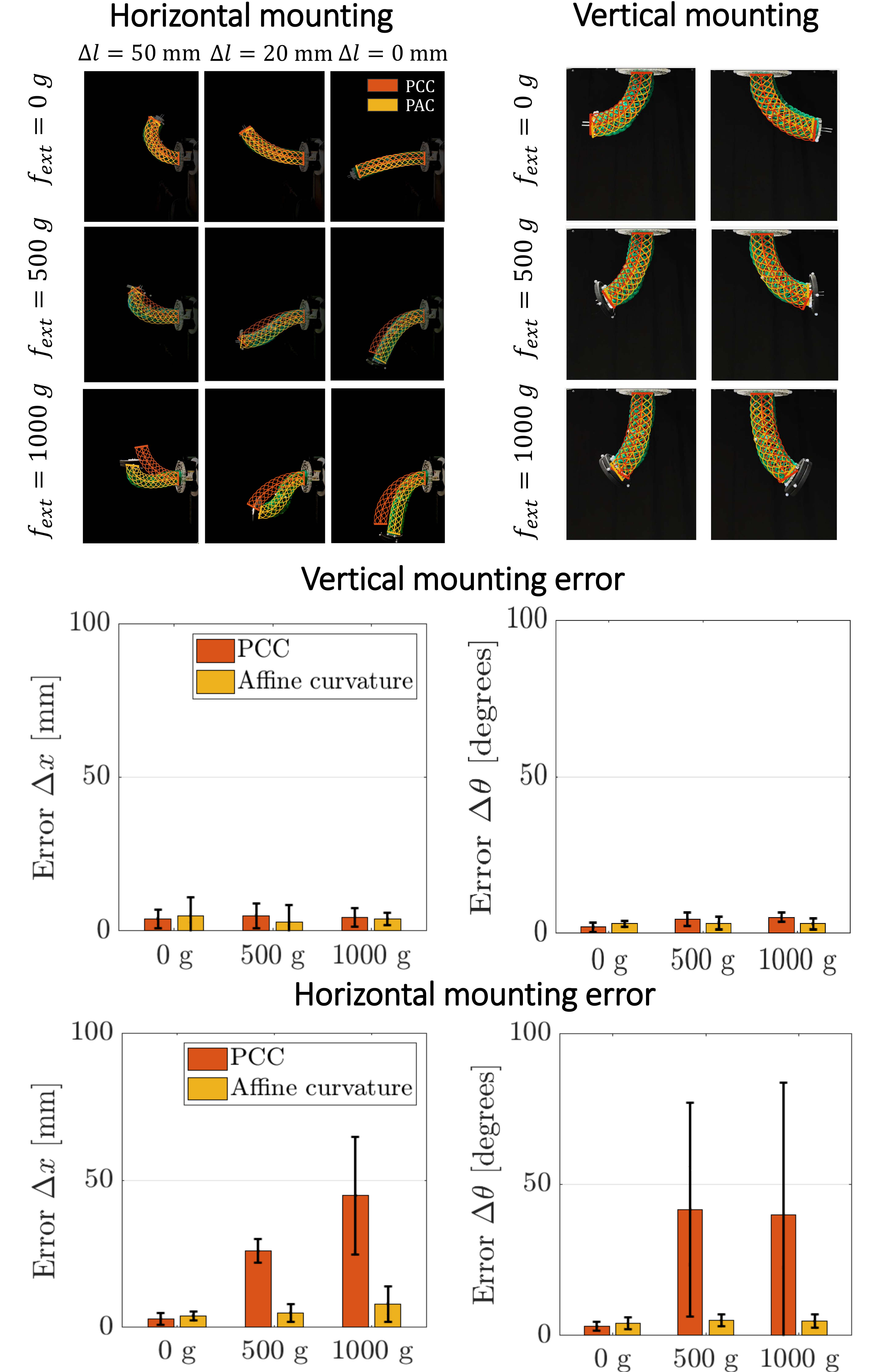}
 \quad
   \caption{Demonstration of the modeling accuracy achieved with different loading conditions with PCC and PAC. On the top-left, the load is directed so to have a significant moment arm, while on the bottom-right, the load is placed so to have a small moment arm. The barcharts report quantitative results on the modeling accuracy obtained with PCC and PAC respectively. }
   \label{fig:single}
\end{figure} 

In order to validate the robustness of the piecewise affine curvature model, we compute the static equilibrium over a variety of loading conditions and imposed tendon lengths. The static equilibrium is solved for both the PAC and PCC models, to allow for comparison of their modeling accuracy. We perform this for different loading conditions and different actuation configurations to explore how the accuracy varies for different conditions. Increasing mass was applied to the end of the single segments  of values $0,500$ or $1000 g$ for three different tendon lengths and for two different configurations of the segment. Fig.~\ref{fig:single} shows an image of the single segment overlaid with the PCC and PAC solutions for each of these experiments. This is shown for varying loads and tendon configurations. We also consider varying directions of the load by varying the placement of the soft structure between a first configuration perpendicular to gravity, and second placement parallel to gravity, where the moment arm is lower.   
The performance of the PAC or PCC model is evaluated by computing the error between the measured section and model in terms of euclidean distance and orientation. The results are reported in the barcharts in Fig.\ref{fig:single}. To compute the latter, we used the Frobenius norm of difference between the end effector orientation and the orientation reconstructed by the models.
The PCC model shows comparable reconstruction performances with respect to PAC for the unloaded case, with a relative average error of $3.2$ mm, $2.1$ degrees as shown in the bottom right part of Fig.\ref{fig:single}. 
\begin{figure*}[t]
 \begin{minipage}{1\textwidth}
 \centering
   \includegraphics[width=1\linewidth]{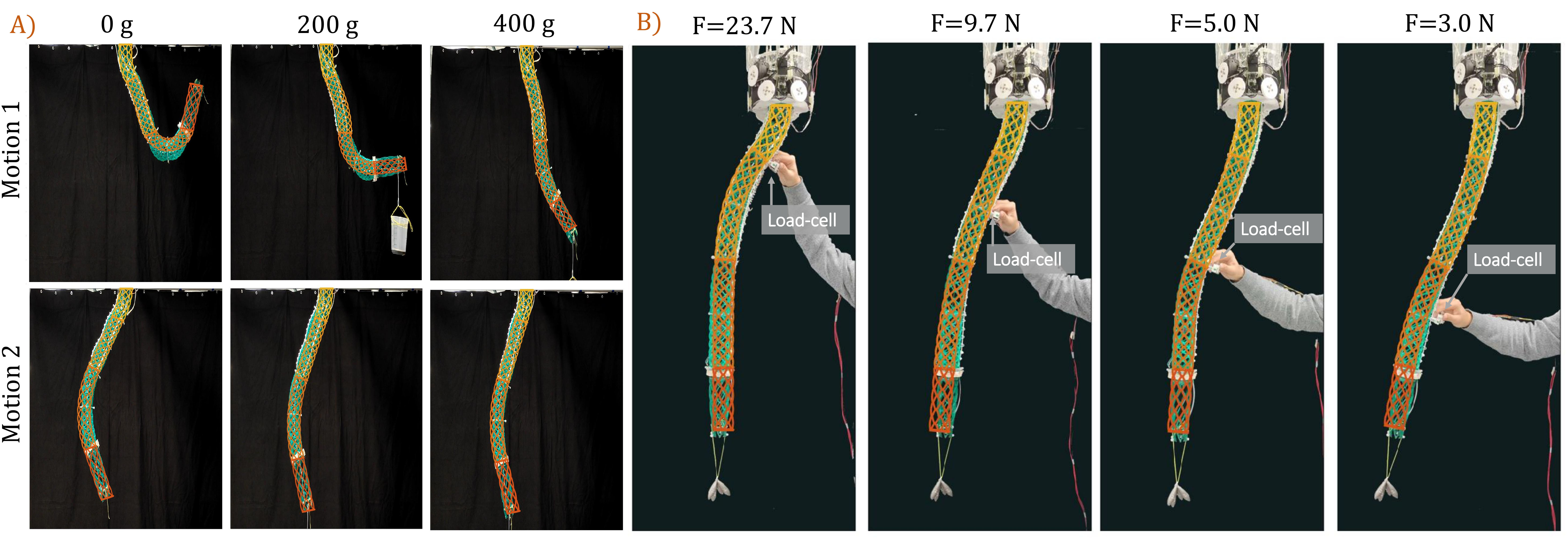}
  \end{minipage} \quad
   \caption{\textbf{A)}: demonstration of the modeling accuracy achieved with different loading conditions, specified at the top, and tendon configurations, specified on the left. \textbf{B)} demonstration of the modeling accuracy achieved when the manipulator is simultaneously perturbed by a load at the tip of $3.41 \mathrm{N}$ and by an external force exerted by a human operator, displayed at the top of the respective figure. }
   \label{fig:full_robot_experiment}
\end{figure*} 
As the external load applied to the structure increase, the curvature of the segments becomes increasingly non-constant and the PAC model significantly outperforms the traditional PCC models. This is particular the case for the perpendicularly mounted segment where the moment arm is far higher, and the curvature of the segment can be far from constant.  Although in some cases the distance error for these can be relatively low, the error in the orientation reflects the significantly different form that is predicted. In the cases with most non uniform curvature, the PAC model can provide a significant error reduction with a relative average error of $27$ mm, $29$ degrees. For the parallel mounted segment the error between the PAC and PCC model is similar, with the segment showing more constant curvature. Thus, the PAC model can significantly outperform the PCC model, and particularly so when high loads or moment arms are applied.

\subsection{Multi-segment reconstruction}

To evaluate the performance of the PAC model on a multi-segment soft robot manipulator, we next consider applying external loads to the full manipulator.
The three segment soft manipulator presented in Figure \ref{fig:Robot design} was loaded with $0,200$ and $400$ g at the tip and two different motion paths are tested. Motion capture markers were placed at the extremes and in the centre of each segment, so to record ground truth data of the manipulator's configuration. 

In Fig.\ref{fig:full_robot_experiment}, it can be observed that the 3D PAC model is able to accurately reconstruct the static equilibrium of the soft manipulator for a variety of  
tendon configurations (e.g. motions) when a range of loads are applied to the end of the manipulator. The PAC model captures the wide range of poses shown by the robot, resulting from the different tendon configurations. These results also highlight how the impact of the load on the pose of the robot is a function of the workspace, and despite this, the PAC model captures the behaviour at different poses within the workspace of the robot.
Moreover, to validate the robustness of the PAC model with respect to environmental interaction, the manipulator is simultaneously perturbed by a human operator whilst a load is also applied at the tip. The force produced by the human interaction is measured with a load-cell (TAL220, $10$ kg max force) which is held by the user. The load-cell is placed perpendicularly to the local curvature of the soft robot. This perpendicular force is then projected on a Lagrangian force on the curvature states through the Jacobian of the contact point, with the methodology presented in \cite{stella2022experimental}. The experiment is repeated 4 times with different locations of interactions whilst the load at the tip is kept constant. Fig.\ref{fig:full_robot_experiment} demonstrates that for the different interaction points, the PAC accurately captures the configuration of the robot when it is subject to external interactions which occur along the length of the robot.


\section{Discussion \& Conclusions}

In this work we demonstrate the need for higher order models to reconstruct the pose of soft robot when interacting with the environment. We introduce a novel model, the 3D piecewise affine curvature model (PAC), and demonstrate experimentally the improved performances achieved with this model when the structure is loaded by external forces. Experimental results show that a significant improvement in modeling accuracy with respect to PCC can be achieved at the cost of only an extra Lagrangian variable per segment. The results show that for larger loads and higher moment arms the pose of the robot differs more significantly from the PCC model predictions and that the novel PAC model is able to greatly improve the reconstruction accuracy. This enables the PAC model to capture the soft robot pose when in contact with humans operators and when carrying varying loads. This novel modelling approach has the potential to significantly improve the control and modelling of soft manipulators by enabling load compensation and the accounting of soft robot-human interactions.  However, the main limitation of the proposed methodology is the lack of a closed solutions for static equilibrium, which makes solving a quasi-static trajectory tracking problem computationally challenging. 

Due to the strong translation of the affine curvature model to real hardware, we believe that a number of results found in simulation for multi-stable soft structures \cite{della2020softinverted} could also be brought to reality. Of particular interest is the theory developed in \cite{trumic2022stability} which exploits the multi-stable  behaviour observed in simulation to generate efficient and guided motion. Deploying such theory on a real hardware could bring to reality the morphological control strategy envisioned by Fuchslin \textit{et al.} \cite{fuchslin2013morphological}, ultimately merging the control policy in the embodiment of the soft robot. In future work, we plan to extend this work for load-aware trajectory planning, morphology based control, as well as using the model as a baseline to perform computational design of fully-fledged soft robotic manipulators.

\bibliographystyle{IEEEtran}
\bibliography{references}

\end{document}